\documentclass[pdflatex,sn-mathphys]{sn-jnl}% Math and Physical Sciences Reference Style

\jyear{2023}%

\theoremstyle{thmstyleone}%
\theoremstyle{thmstyletwo}%

\theoremstyle{thmstylethree}%
\makeatletter \def\ps@pprintTitle{  \let\@oddhead\@empty  \let\@evenhead\@empty  \def\@oddfoot{\hfill\thepage}  \def\@evenfoot{\thepage\hfill}} \makeatother

\begin{document}

\title [Accepted on Jan 19, 2024 in the Cybersecurity, Springer Open Journal] {Securing Transactions: A Hybrid Dependable Ensemble Machine Learning Model using IHT-LR and Grid Search}

\author*[1]{Md. Alamin Talukder} \email{alamin.cse@iubat.edu}
\author*[2]{Rakib Hossen} \email{rakib0001@bdu.ac.bd}
\author[3]{Md Ashraf Uddin} \email{ashraf.uddin@deakin.edu.au}
\author[4]{Mohammed Nasir Uddin} \email{nasir@cse.jnu.ac.bd}
\author[4]{Uzzal Kumar Acharjee	} \email{uzzal@cse.jnu.ac.bd}

\affil[1]{Department of Computer Science and Engineering, International University of Business Agriculture and Technology, Dhaka, Bangladesh}
\affil[2]{Bangabandhu Sheikh Mujibur Rahman Digital University, Bangladesh Kaliakoir, Gazipur, 1750, Bangladesh}
\affil[3]{School of Information Technology, Deakin University, Waurn Ponds Campus, Geelong, Australia}
\affil[4]{Department of Computer Science and Engineering, Jagannath University, Dhaka, Bangladesh}

\abstract{
Financial institutions and businesses face an ongoing challenge from fraudulent transactions, prompting the need for effective detection methods. Detecting credit card fraud is crucial for identifying and preventing unauthorized transactions. While credit card fraud incidents are relatively rare, they can result in substantial financial losses, particularly due to the high monetary value associated with fraudulent transactions. Timely detection of fraud enables investigators to take swift actions to mitigate further losses. However, the investigation process is often time-consuming, limiting the number of alerts that can be thoroughly examined each day. Therefore, the primary objective of a fraud detection model is to provide accurate alerts while minimizing false alarms and missed fraud cases. In this paper, we introduce a state-of-the-art hybrid ensemble (ENS) dependable Machine learning (ML) model that intelligently combines multiple algorithms with proper weighted optimization using Grid search, including Decision Tree (DT), Random Forest (RF), K-Nearest Neighbor (KNN), and Multilayer Perceptron (MLP), to enhance fraud identification. To address the data imbalance issue, we employ the Instant Hardness Threshold (IHT) technique in conjunction with Logistic Regression (LR), surpassing conventional approaches. Our experiments are conducted on a publicly available credit card dataset comprising 284,807 transactions. The proposed model achieves impressive accuracy rates of 99.66\%, 99.73\%, 98.56\%, and 99.79\%, and a perfect 100\% for the DT, RF, KNN, MLP and ENS models, respectively. The hybrid ensemble model outperforms existing works, establishing a new benchmark for detecting fraudulent transactions in high-frequency scenarios. The results highlight the effectiveness and reliability of our approach, demonstrating superior performance metrics and showcasing its exceptional potential for real-world fraud detection applications.
}


\maketitle

\section{Introduction}
\label{introduction}

Fraudulent transactions refer to unauthorized or deceptive activities conducted to obtain financial gains illegally. These transactions can take various forms, including credit card fraud, identity theft, money laundering, and insurance scams  \citep{raphael2023card, kartheek2023analysis}. Fraudulent activities have become a major concern for businesses and organizations worldwide, leading to significant financial losses and damage to reputations \citep{kayode2023applications}. A comprehensive study by the Association of Certified Fraud Examiners (ACFE) highlights the significant financial impact of fraud, estimating annual global losses at over \$4.7 trillion. This analysis, based on 2,110 cases across 133 countries, revealed staggering losses of \$3.6 billion, averaging \$1.78 million per case. Compiled by a network of over 90,000 anti-fraud professionals, the data underscores fraud's ubiquity across various industries \citep{acfe2022}. Detecting fraudulent credit card transactions is crucial for businesses to surpass other financial crimes in scale and cost, highlighting the COVID-19 pandemic and rising cryptocurrency fraud as new challenges, with organizations losing about 5\% of their annual revenue to these issues and financial institutions to protect themselves and their customers \citep{acfe2022global, faccia2023national}. It not only helps prevent financial losses but also safeguards the integrity of the entire financial system. Existing methods of fraud detection, relying on manual inspection, rule-based systems and AI-based automated models, have proven to be inadequate in detecting sophisticated and evolving fraudulent activities \citep{al2023role, alfaiz2022enhanced, taha2020intelligent, lakshmi2018machine}. Hence, there is a growing need for more advanced and automated approaches to detect and prevent fraudulent transactions. 

Machine learning (ML) techniques have emerged as powerful tools for fraud detection due to their ability to analyze large volumes of data and identify complex patterns \citep{talukder2023dependable, talukder2023efficient, talukder2023empowering}. ML algorithms can learn from historical transaction data and detect anomalies or patterns indicative of fraudulent activities \citep{talukder2024machine, talukder2024mlstlwsn}. They offer several advantages: scalability, efficiency, and adaptability to changing fraud patterns. By leveraging ML algorithms, organizations can significantly improve their fraud detection capabilities. Ensemble learning (EL) is an ML technique that combines the predictions of multiple individual models to make a final prediction \citep{talukder2022machine}. It has shown great promise in various domains, including fraud detection. Ensemble methods can reduce overfitting, enhance model generalization, and improve overall performance by leveraging the strengths of different models \citep{chang2022digital}. Combining multiple models with optimization allows EL to capture diverse patterns and characteristics of fraudulent transactions, leading to more accurate and reliable detection.

Several existing works have addressed fraudulent transaction detection using ML and EL techniques. For instance, Esenogho et al. applied Synthetic Minority Over-sampling Technique-Edited Nearest Neighbors (SMOTE-ENN) with Long Short-Term Memory (LSTM) for identifying credit card fraud \citep{esenogho2022neural}. Dornadula et al. used SMOTE data balancing techniques and random forests for classifying credit card transactions \citep{dornadula2019credit}. Lakshmi et al. employed ML models, including Logistic Regression, Decision Tree, and Random Forest for credit card fraud detection \citep{lakshmi2018machine}. Xie et al. proposed a Heterogeneous EL Model based on Data Distribution (HELMDD) to handle the data imbalance issue in detecting credit card fraud \citep{xie2021heterogeneous}. However, these existing works have certain limitations such as limited improvement in accuracy and failure to address the highly imbalanced nature of fraudulent transaction data.

This paper presents a hybrid ensemble ML model for fraudulent transaction detection. Our approach aims to address the limitations of existing works and improve the accuracy of fraud detection. The hybrid ensemble model combines multiple ML algorithms, including Decision Tree (DT), Random Forest (RF), K-Nearest Neighbour (KNN) and Multilayer Perceptron (MLP) with optimization to leverage their complementary strengths and capture diverse patterns in the data. We employ preprocessing techniques such as standardization, label encoding, and resampling using Instance hardness threshold with Logistic Regression (IHT-LR) to address data quality and imbalance challenges. We demonstrate our proposed model's high accuracy and generalizability through extensive experimentation on a real-world credit card fraud dataset.

% By combining these algorithms, we aim to capture diverse patterns and characteristics of fraudulent transactions, leading to improved performance compared to individual models or traditional approaches.
The key contributions of our research are as follows:
\begin{itemize}

\item Introduced a novel hybrid ensemble dependable ML model specifically designed for credit card fraudulent transaction detection. It integrates extensive preprocessing with data balancing using IHT-LR and combining multiple ML algorithms with proper optimization, leveraging their strengths and enhancing fraud detection.

\item Conducted an extensive and rigorous assessment of the performance of our proposed model, including various performance metrics evaluation, dependable analysis, complexity analysis and comparison with existing works in credit card fraudulent transactions. Through this comparison, we highlight the superiority and effectiveness of our hybrid ensemble model.
\end{itemize}

Our research significantly contributes to fraudulent transaction detection by addressing the limitations of existing approaches and incorporating state-of-the-art techniques. The proposed hybrid ensemble model improves the accuracy and reliability of fraud detection, thereby enhancing the security and trustworthiness of financial systems. The performance analysis and comparison demonstrate our approach's practical applicability and effectiveness in real-world scenarios. We anticipate that our research will greatly impact the security field and contribute to developing robust fraud detection systems, benefiting businesses, organizations, and individuals alike.

The remaining sections of this paper are structured as follows: In Section \ref{sec:related}, we present a comprehensive review of related literature focusing on credit card fraudulent transaction detection. Section \ref{sec:method} provides a detailed explanation of our research methodology and includes a description of the dataset used. The experimental setup and performance evaluation are outlined in Section \ref{sec:result}. In section \ref{sec:discuss} and \ref{sec:complex}, we presented the resutls discussion and complexity analasis of our research. Furthermore, in Section \ref{sec:depend}, we conduct a thorough analysis of the dependability of our proposed approach. Lastly, Section \ref{sec:conclusion} presents the conclusion and future work.

\section{Related Works}
\label{sec:related}
Many additional studies have explored using hybrid ensemble machine-learning models for fraudulent transaction detection. 

\citep{esenogho2022neural} introduced an innovative credit card fraud detection approach that utilized a neural network ensemble (NNE) and a hybrid data resampling method. Long Short-Term Memory (LSTM) was employed as the base learner in an Adaptive Boosting (AdaBoost) framework, and Synthetic Minority Oversampling Technique-Edited Nearest Neighbors (SMOTE-ENN) was used for data resampling. The proposed method outperformed benchmark algorithms, achieving a sensitivity of 99.60\% and specificity of 99.80\%. The findings showcased the approach's effectiveness in detecting credit card fraud and offered the potential for enhanced security in real-world transactions.

\citep{jovanovic2022tuning} presented an innovative approach for credit card fraud detection that combined ML with a swarm metaheuristic, namely the Group Search Firefly (GSF) algorithm. The enhanced algorithm was used to fine-tune popular machine-learning models such as Support Vector Machine (SVM), Extreme Learning Machine (ELM), and eXtreme Gradient Boosting (XGBoost). Experiments were conducted on a real-world dataset from European credit card transactions, and the results demonstrated that the proposed approach outperformed other state-of-the-art methods. The findings indicated that the GSF algorithm could enhance credit card fraud detection, making it an attractive option for researchers and practitioners.

\citep{soleymanzadeh2022cyberattack} proposed an ensemble stacking method for efficient detection of cyberattacks in the Internet of Things (IoT). The authors conducted experiments on three datasets, including credit cards, NSL-KDD, and UNSW. Their stacked ensemble classifier outperformed individual base model classifiers, achieving an impressive accuracy rate of 93.49\%. This breakthrough demonstrates the potential of their approach in effectively addressing cyberattacks and credit card fraud.
\citep{nandi2022credit} presented a cutting-edge approach to credit card fraud detection using a multi-classifier framework. The ensemble model combines multiple ML classification algorithms, leveraging the Behavior-Knowledge Space (BKS) to combine their predictions. Real-world financial records and publicly available datasets were used for performance evaluations, and the BKS-based ensemble model achieved an impressive accuracy rate of 99.81\%. Statistical tests confirmed the superiority of the developed model compared to commonly used methods like majority voting for noisy data classification and credit card fraud detection. The findings highlight the effectiveness of the proposed approach in tackling these challenges, making it a compelling choice for researchers and practitioners in the field.

Detecting fraudulent activities is of utmost importance for financial institutions. Therefore, it is crucial to employ a fast and efficient model for data handling. In this regard, \citep{faraji2022review} conducted a thorough evaluation of different techniques using real-world data and proposed an ensemble model as a potential solution. They combined XGBoost with SMOTE to address the data imbalance issue, leading to an impressive accuracy rate of 99\%. This study highlights the significance of utilizing ensemble models to effectively handle imbalanced data and achieve high accuracy in credit card fraud detection.

\citep{alfaiz2022enhanced} analyzed 66 ML models in two stages to detect credit card fraud using a dataset of European cardholders. In the first stage, nine algorithms were tested, and the top three were selected for the second stage, along with 19 resampling techniques. After evaluating 330 metric values, the AllKNN undersampling technique combined with CatBoost (AllKNN-CatBoost) was the best model with an Area Under Curve (AUC) value of 97.94\%. This proposed model outperformed previous models and was compared with related works.

\citep{taha2020intelligent} presented a novel approach for detecting fraud in credit card transactions using an optimized light gradient boosting machine (OLightGBM). The proposed approach integrates a Bayesian-based hyperparameter optimization algorithm to tune the parameters of the light gradient boosting machine (LightGBM) intelligently. Two real-world public credit card transaction datasets comprising fraudulent and legitimate transactions were used for experiments. Results showed that the proposed OLightGBM approach outperformed other methods, achieving the highest accuracy (98.40\%) compared to other approaches.

\citep{xie2021heterogeneous} presented a novel approach called the Heterogeneous EL Model based on Data Distribution (HELMDD) to address the challenge of imbalanced data in Credit Card Fraud Detection (CCFD). Through comprehensive experiments on actual credit card datasets, HELMDD demonstrated superior performance compared to existing state-of-the-art models. HELMDD achieved high recall rates for minority and majority classes while significantly increasing banks' savings rates to 86.23 and 66.96, respectively. These findings suggest that HELMDD is an effective solution for addressing imbalanced data in CCFD and can potentially benefit financial institutions in detecting credit card fraud.

In light of increasing fraud rates, \citep{dornadula2019credit} aimed to develop a novel fraud detection method using ML techniques. They utilized a real-world dataset of European credit card frauds to analyze past transaction details and extract behavioral patterns of cardholders. By employing advanced techniques such as RF and synthetic minority oversampling technique (SMOTE), our approach achieved an impressive accuracy rate of 99.98\%. Our research represents a significant advancement in credit card fraud detection, showcasing the potential of our method to effectively detect and analyze frauds in online transactions, protecting consumers and businesses from financial losses.

\citep{lakshmi2018machine} examined the performance of LR, DT, and RF for credit card fraud detection using a dataset of credit card transactions from a European bank. The dataset was highly imbalanced, with a low percentage of fraudulent transactions; oversampling was done to balance the dataset. The results showed that the accuracy of LR, DT, and RF was 90.0\%, 94.3\%, and 95.5\%, respectively. Comparative analysis revealed that RF outperformed LR and DT techniques.

\citep{kalid2020multiple} employed a Multiple Classifiers System (MCS) to analyze two datasets: credit card frauds (CCF) and credit card default payments (CCDP). The MCS utilized a sequential decision combination strategy to achieve accurate anomaly detection. The study found that the MCS outperformed existing research, particularly in detecting minority anomalies in these credit card datasets. Overall, the proposed MCS exhibited superior performance in handling credit datasets with overlapping classes and unbalanced class distribution.

\section{Methodology}
\label{sec:method}

%\color{blue}
We introduced a novel hybrid ensemble model that intelligently combines these algorithms, leveraging their complementary strengths to enhance fraud detection accuracy significantly. Our approach integrates advanced techniques such as Instance Hardness Threshold with Logistic Regression (IHT-LR) to address data imbalance issues, ensuring the robustness of our model. Furthermore, we employed grid search optimization to fine-tune the parameters of our ensemble model, maximizing its performance. By systematically exploring the parameter space, grid search enables us to identify the most effective combination of algorithms and their respective weights. Therefore, while the individual algorithms may be popular, our methodology represents a novel and effective approach to credit card fraud detection, offering a valuable contribution to the field.
%\color{black}

In this section, we present the methodology employed in our study for detecting fraudulent transactions. We outline the main stages of our proposed methodology, including data collection, pre-processing, normalization, balancing, splitting, ensemble model building, and performance analysis. The objectives of our research are highlighted in the following sections:

\subsection{Research Objectives}

\begin{enumerate}
\item \textbf{Data Preprocessing Techniques:} The objective of this research is to investigate and apply effective data preprocessing techniques, including standardization and label encoding, to enhance data quality and address the imbalanced nature of fraudulent transaction datasets using the Instance Hardness Threshold (IHT) resampling method.

\item \textbf{Hybrid Ensemble ML Model:} The primary objective is to develop a novel hybrid ensemble ML model specifically designed for detecting credit card fraudulent transactions. This model will intelligently integrate Decision Tree (DT), Random Forest (RF), K-Nearest Neighbor (KNN), and Multilayer Perceptron (MLP) algorithms, leveraging their strengths to improve fraud detection accuracy significantly.

\item \textbf{Performance Analysis and Comparison:} The research aims to conduct a comprehensive performance analysis of the proposed hybrid ensemble ML model using a real-world credit card fraud dataset. Furthermore, a rigorous comparison with existing approaches in fraudulent transaction detection will be performed to demonstrate the superior performance and effectiveness of the proposed model.

\item \textbf{Contribution to the Field:} The research objective is to make a valuable contribution to the field of fraudulent transaction detection by addressing the limitations of existing approaches and incorporating state-of-the-art techniques. The proposed hybrid ensemble ML model's improved accuracy and reliability in detecting fraud will enhance the security and trustworthiness of financial systems, benefiting businesses, organizations, and individuals.
\end{enumerate}

Figure \ref{fig:proposal} illustrates our proposed hybrid dependable machine learning model for fraudulent transaction detection. The proposed developed methodology for the fraudulent transaction detection model involved the following steps:

\begin{figure*}[!htbp]
	\centering
	{\includegraphics[scale=.195]{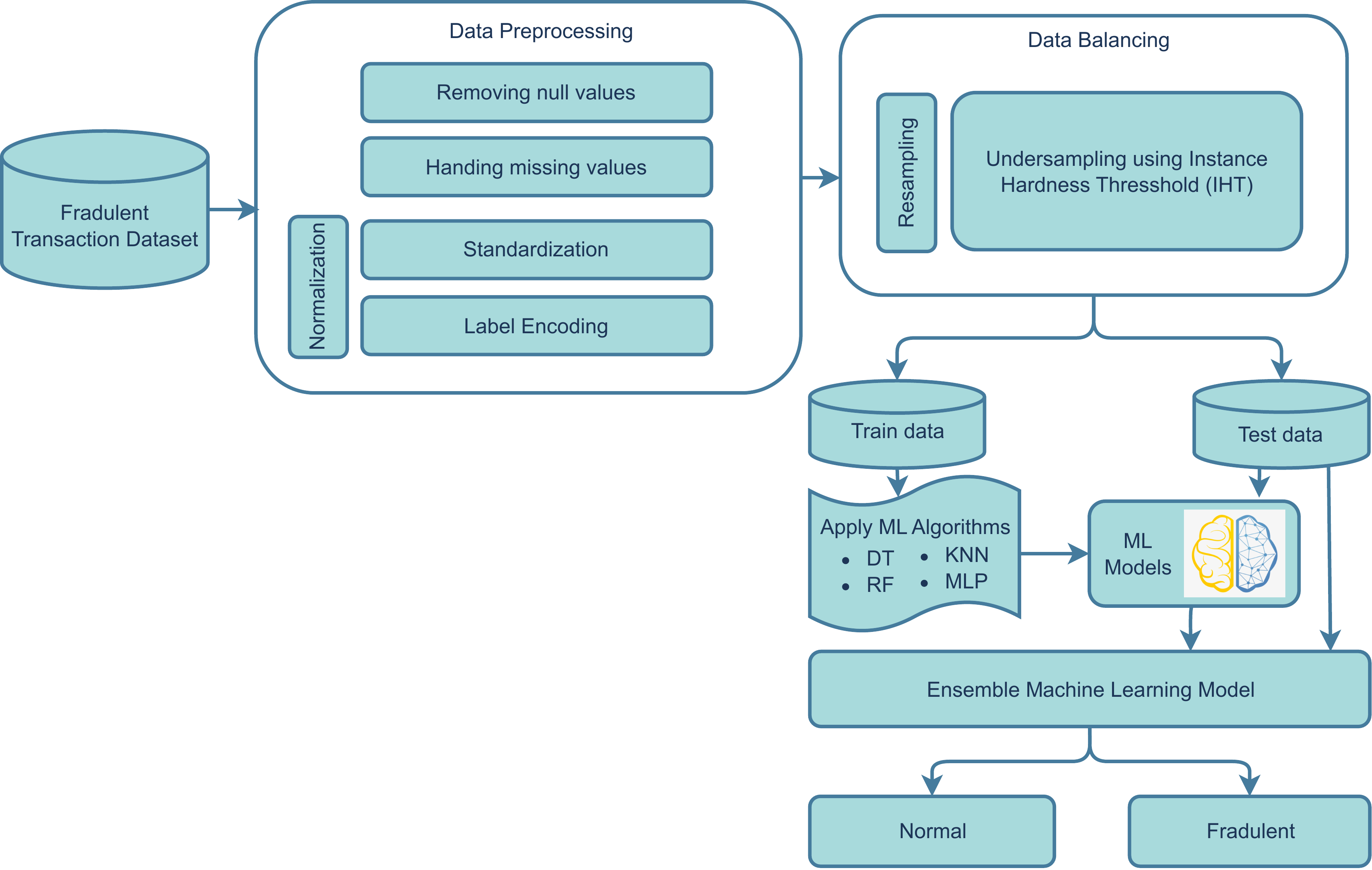}}
	\caption{The proposed fraudulent transaction detection architecture}
	\label{fig:proposal}
\end{figure*}

\begin{itemize}
\item Dataset Selection: We obtained a credit card fraudulent transaction dataset as our input dataset. This dataset served as the foundation for training our fraudulent transaction detection model.
\item Data Preprocessing: In this stage, we performed data preprocessing tasks to ensure the quality and suitability of the dataset. We addressed missing values by removing null values and handling any missing values appropriately. Additionally, we applied normalization techniques, such as standardization and label encoding, to transform the data into a normalized form suitable for machine learning models.
\item Handling Data Imbalance: To tackle the issue of imbalanced data, we employed the Instant Hardness Threshold (IHT) technique in conjunction with the Logistic Regression (LR) algorithm. This approach helped us address data imbalance, ensuring that our dataset was balanced and avoiding the problems of overfitting and underfitting.
\item Dataset Splitting: We partitioned our dataset into two parts using the k-fold cross-validation process. This splitting technique allowed us to create separate training and test sets. The training set was utilized for model building, while the test set served for evaluating the model's predictive performance.
\item Ensemble Model Initialization: To create our ensemble model, we applied multiple machine learning algorithms, including Decision Trees (DT), Random Forests (RF), k-nearest Neighbors (KNN), and Multilayer Perceptron (MLP). Each algorithm contributed its unique capabilities to the ensemble model.
\item Ensemble Model Construction: We combined the predictions from the individual machine learning algorithms to construct our ensemble model. By leveraging the collective intelligence of multiple models, the ensemble approach enhanced the accuracy and reliability of fraudulent transaction detection.
\item Performance Analysis: Finally, we conducted an extensive performance analysis of our proposed model. This analysis involved evaluating various performance metrics, such as accuracy, precision, recall, F1 score and dependability analysis, etc, to assess the effectiveness of our model in identifying fraudulent transactions accurately.
\end{itemize}

By following this methodology, we aimed to develop a robust and effective fraudulent transaction detection model. Our approach included careful data preprocessing, addressing data imbalance, utilizing ensemble modeling techniques, and conducting comprehensive performance analysis to validate the model's effectiveness in identifying fraudulent transactions.

\subsection{Dataset Description}

The Credit Card Fraud Detection dataset is a public dataset available on Kaggle \citep{creditcardfraud} that contains anonymized credit card transactions made over two days in September 2013 by European cardholders. The dataset contains a total of 284,807 transactions, of which 492 are fraudulent. The dataset has 28 features, of which 27 are numerical features generated by Principle component Analysis (PCA) transformation due to confidentiality reasons, and the last feature is the class label indicating whether the transaction is fraudulent or not. The dataset is highly imbalanced, with the majority of transactions being non-fraudulent. This poses a challenge for building a classifier that can accurately detect fraudulent transactions while minimizing false positives. The dataset has been used in several research studies and competitions to develop and evaluate ML models for credit card fraud detection. The dataset is intended for use in the research and development of fraud detection algorithms and is not intended for commercial use. The dataset is publicly available and can be used for educational or research purposes. However, the use of this dataset for any commercial purposes requires the prior written consent of the dataset owners.

While acknowledging the significant changes in the payment landscape over the past decade, this dataset was chosen for its comprehensive nature and the depth of transactional data it provides, which is rare in more recent datasets due to privacy concerns. The historical data allows us to trace the evolution of fraud tactics, offering insights into how fraudulent behaviors have adapted over time. Additionally, the fundamental principles of fraud detection extracted from this dataset can be applied to newer systems, offering a foundational understanding that remains relevant. Lastly, the dataset serves as a benchmark, allowing us to compare the effectiveness of modern fraud detection techniques against historical data.

\subsection{Data Preprocessing}
In the data processing section of our study, we meticulously handled missing values and null values in the fraudulent transaction dataset to ensure integrity and reliability. Our systematic approach, including careful imputation techniques and thorough documentation of decisions, created a clean and robust dataset for analysis. By effectively addressing potential biases and uncertainties, we maintained transparency and reproducibility, resulting in findings based on a reliable dataset.

In the data normalization, we employed standardization and label encoding techniques to ensure that the data was suitable for ML models. Standardization involved scaling the data to have zero mean and unit variance, while label encoding was used to convert categorical variables into numerical representations. These normalization techniques helped to align the data and bring it to a consistent and comparable scale, allowing for effective training and evaluation of ML models.

\textbf{Standardization:} Standardization transforms the data to have zero mean and unit variance, making it centered around 0 and scaled to have a standard deviation of 1.

The formula for standardization is:
\begin{equation}
x_{\text{standardized}} = \frac{{x - \text{mean}(x)}}{{\text{standard deviation}(x)}}    
\end{equation}
where: {x} represents the original data point
{mean(x)} is the mean of the data
{standard deviation(x)} is the standard deviation of the data

\textbf{Label Encoding:} Label encoding enables ML models to process categorical data as numerical inputs, allowing for mathematical computations and analysis. The simplest form of label encoding is assigning integer labels to each unique category in the data. 

The formula for label encoding is:
\begin{equation}
x_{\text{encoded}} = \text{Label}(x)    
\end{equation}

where: {x} represents the original categorical variable and 
{Label(x)} is the numerical label assigned to each unique categorical value. For instance, we have normal and fraudulent labels; this Label encoding converts this into 0 and 1 for normal and fraudulent.

\subsection{Data Balancing}

In the data balancing step of our study, we employed resampling techniques to address the class imbalance issue in our dataset. In combination with the instance hardness threshold (IHT), undersampling is a crucial step in addressing the class imbalance issue in credit card fraud detection for financial data analysis. So, we utilized undersampling and applied an IHT using LR to achieve a better-balanced dataset. In this process, undersampling involves randomly removing instances from the majority class to create a balanced dataset, while IHT with LR allows us to select instances from the majority class that are more likely to be misclassified. This approach allowed us to balance the classes without generating synthetic data or losing any information while also ensuring that we utilized the original data for training and testing our model. We built a Credit Card Fraud Transaction (CCFT) detection model using this balanced dataset, aiming for real-world applications in fraud detection, improving its accuracy and reliability.

\subsection{Applied ML Models}

We have also utilized four other popular ML algorithms for credit card fraudulent detection: DT, RF, KNN, and MLP.

\begin{itemize}
    \item \textit{Decision Tree (DT):} DT is a flexible classifier that performs well in both classification and regression tasks, making it a valuable tool for data analysis and prediction \citep{DT}. Its ability to capture relevant decision-making information from the dataset and represent it in tree structures makes it interpretable and robust. Due to its wide applicability and interpretability, DT is highly regarded by researchers and practitioners in various fields.

    \item \textit{Random Forest (RF):} RF, an ensemble classifier, is a powerful technique that combines the collective intelligence of multiple DTs to improve accuracy and generalization \citep{RF}. Its capability to handle high-dimensional data and mitigate overfitting makes it an attractive option for complex real-world problems. With its scalability and accuracy, RF is widely adopted in various domains, including finance, healthcare, and natural language processing, among others.

    \item \textit{K-Nearest Neighbor (KNN):} KNN, a cutting-edge supervised learning method, is a powerful tool for data classification \citep{KNN}. Its simplicity and efficiency make it popular among researchers and practitioners for its ability to accurately classify data points based on their proximity to existing categories in the training dataset. With its interpretable approach and high accuracy, KNN is widely used in fields such as pattern recognition, image processing, and recommendation systems \citep{uddin2023ensemble}.

    \item \textit{Multilayer Perceptron (MLP):} MLP, a popular type of Artificial Neural Network (ANN), is a versatile tool for complex data processing and analysis \citep{castro2017multilayer}. With its interconnected layers of neurons, activation functions, and hidden layers, MLP can model complex relationships in data and make accurate predictions. Its wide applicability in areas such as image recognition, speech processing, and bioinformatics, among others, makes MLP a valuable asset for researchers and practitioners in diverse fields of study \citep{khatun2023cancer}.
\end{itemize}

Imagine a powerful credit card fraud detection system that utilizes not just one but four cutting-edge algorithms working together in a hybrid ensemble! With DT, RF, KNN and MLP at its disposal, this system is designed to catch even the most sophisticated fraudsters. By evaluating their performance using metrics like precision, recall and accuracy, etc, we can ensure that this dynamic combination of algorithms is truly unbeatable in identifying fraudulent transactions and safeguarding your financial security.

\subsection{Ensemble ML Approach}

In our study, we employed an ensemble ML approach to improve the accuracy and robustness of our credit card fraud detection model. Specifically, we utilized a weighted average voting ensemble approach using grid search techniques, combining the predictions from multiple ML models such as DT, RF, KNN and MLP.

\begin{figure*}[!htbp]
	\centering
	{\includegraphics[scale=.25]{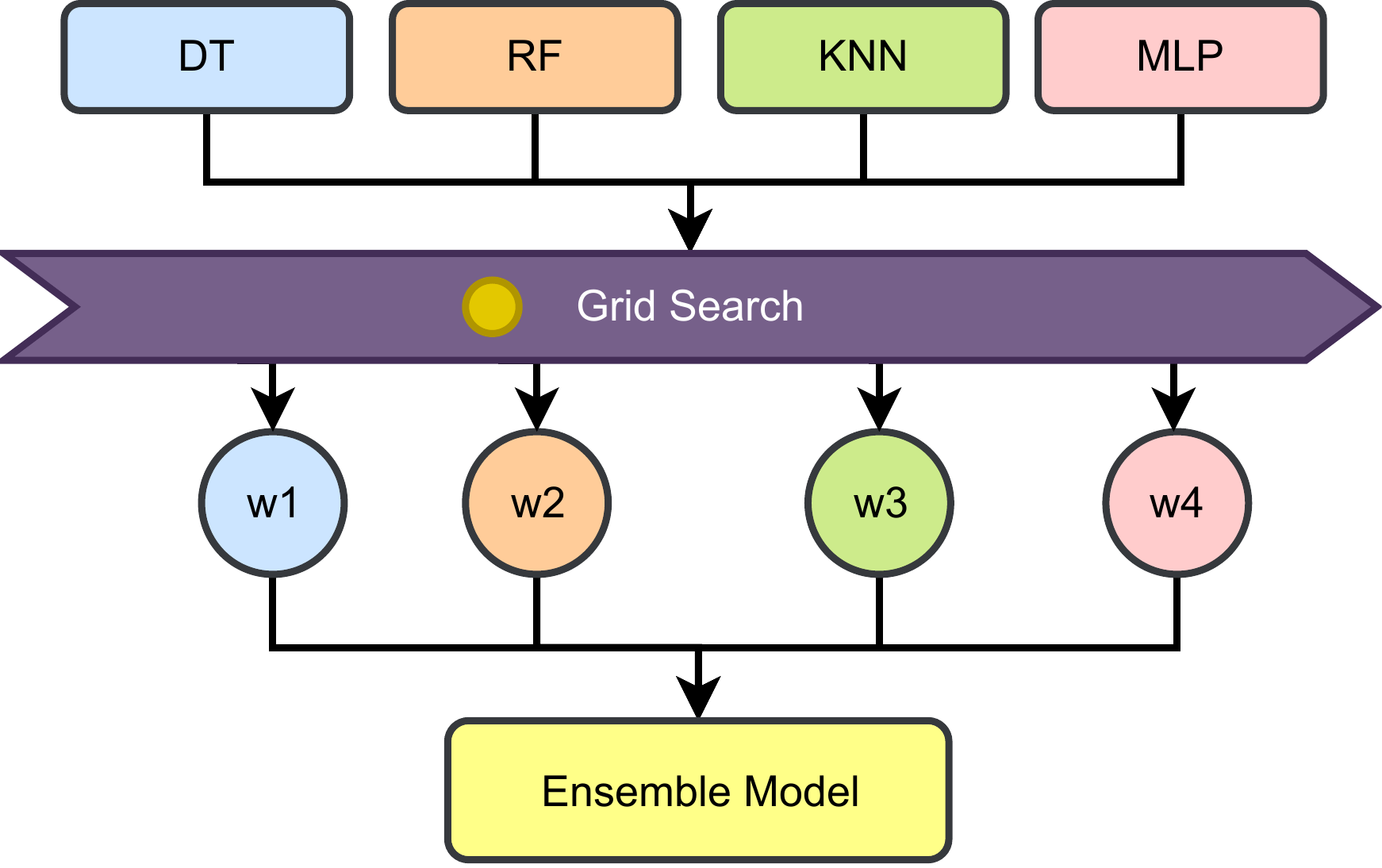}}
	\caption{The proposed hybrid ensemble approach using Grid Search}
	\label{fig:ens}
\end{figure*}

To assign appropriate weights to each algorithm, we performed a grid search to identify the optimal weights that would result in the best performance of our ensemble model. This process involved systematically tuning the weights for each algorithm to achieve a balanced and accurate ensemble prediction. By utilizing this ensemble approach and optimizing the weights through grid search, we aimed to build a more effective credit card fraud detection model, which could help save the nation from economic loss caused by fraudulent transactions. The Fig. \ref{fig:ens} illustrates the proposed hybrid ensemble approach for CCFT detection.

The weighted average voting ensemble model takes a weighted average of the predictions from each base model, where the weights reflect their relative importance. This approach leverages the strengths of each base model, mitigates their weaknesses, and enhances the overall prediction accuracy. Combining the predictions of multiple base models makes the ensemble model less likely to overfit and more robust in detecting credit card fraud. Using the weighted average voting ensemble model is significant in our study as it improves the performance of the credit card fraud detection model by leveraging the expertise of multiple base models. The optimized weights found from the grid search technique, such as 0.25, 0.5, 0.5 and 0.25 for DT, RF, KNN and MLP, respectively, obtained through the grid search process ensure that the ensemble model is well-calibrated and perform optimally in classifying fraudulent transactions. This can help financial institutions save potential economic losses caused by fraud. The effectiveness and robustness of the weighted average voting ensemble model make it a valuable technique in credit card fraud detection and other ML applications.

 % [0.25, 0.5, 0.5, 0.25] got 100% accuracy rate

%\color{blue}
 \subsection{Exploring Ensemble Weights}

While our study primarily focused on the weighted average voting ensemble approach with optimized weights obtained through grid search, it is essential to consider alternative approaches and discuss their potential implications on the ensemble model's performance.

One alternative approach to ensemble weight selection involves heuristic methods, such as equal weighting or assigning weights based on the performance of individual base models on a validation set. Equal weighting, where each base model is assigned the same weight, provides a straightforward approach but may not fully leverage the varying strengths of each model. Conversely, assigning weights based on individual model performance allows for a dynamic weighting scheme that adapts to the performance of each base model. However, this method requires careful selection criteria and may not always result in optimal ensemble performance.

Furthermore, ensemble weight selection can be influenced by domain knowledge and expert insights. Expertise in specific types of fraudulent transactions may guide the assignment of weights to emphasize the contributions of certain base models known to perform well in relevant scenarios. This approach leverages domain expertise to tailor the ensemble model to the specific challenges of fraud detection.

Additionally, ensemble weight sensitivity analysis can provide valuable insights into the robustness of the ensemble model to variations in weight assignments. By systematically varying the weights and observing changes in performance metrics, researchers can gain a deeper understanding of how different weight configurations affect the ensemble model's predictive capabilities. Sensitivity analysis helps identify optimal weight ranges and potential trade-offs between model accuracy and computational efficiency.

While exploring alternative approaches to ensemble weight selection is valuable, it is essential to highlight the advantages of grid search optimization. Grid search systematically evaluates combinations of weights across a predefined search space, allowing for a comprehensive exploration of weight configurations. This approach enables researchers to identify the optimal combination of weights that maximize the ensemble model's performance metrics, such as accuracy, precision, recall, and F1-score. By searching the entire weight space, grid search minimizes the risk of overlooking promising weight configurations and ensures that the ensemble model is well-calibrated and robust.

%\color{black}

\section{Results and Discussion}
\label{sec:result}
%\color{blue}
In our experiment, we have rigorously evaluated the performance of our hybrid ensemble model for credit card fraud detection, showcasing its robustness and effectiveness in real-world scenarios.
%\color{black}
\subsection{Experiment Setup}
The experimental setup involved conducting the experiments on a high-performance machine equipped with 8 cores, 64 GB RAM, and a 100 GB disk. We leveraged Anaconda Navigator and its Jupyter Notebook interface to implement our proposed model. Python was the primary programming language, accompanied by widely used libraries such as Pandas, NumPy, Matplotlib, Seaborn, TensorFlow, Keras, and Scikit-learn. These libraries facilitated various tasks such as data manipulation, numerical computations, visualization, and machine learning operations. The compressed environment provided by Anaconda Navigator and Jupyter Notebook played a pivotal role in efficiently validating our approach to credit card fraudulent transaction detection. The utilization of these tools, along with the Python libraries, greatly contributed to the success and reliability of our research, enabling us to derive meaningful insights and make advancements in CCFT detection.

%\color{blue}
\subsection{Performance Evaluation Metrics}

In this section, we present the performance evaluation metrics utilized to assess the effectiveness of our proposed model in detecting fraudulent transactions. The following metrics are employed:

\begin{itemize}
\item \textbf{Confusion Matrix}: This matrix provides a comprehensive view of the model's classification performance, categorizing predictions into true positives (TP), true negatives (TN), false positives (FP), and false negatives (FN). Table \ref{table:confusion} shows the confusion matrix.

\begin{table}[!htbp]
    \centering
    \begin{tabular}{lll}
    \hline
        & Actual   positive & Actual   negative \\ \hline
        Predicted   positive & TP & FP \\ 
        Predicted   negative & FN & TN \\ \hline
    \end{tabular}
    \caption{Confusion Matrix}
    \label{table:confusion}
\end{table}

\item \textbf{Accuracy ($Acc$)}: Defined as the ratio of correctly classified instances to the total number of instances, accuracy measures the overall effectiveness of the model in correctly identifying both fraudulent and non-fraudulent transactions.
    \begin{equation}
    Accuracy =\frac{TP+TN}{TP+FP+FN+TN}
    \end{equation}

\item \textbf{Precision ($Prec$)}: Precision quantifies the proportion of correctly identified fraudulent transactions among all transactions classified as fraudulent by the model, thus indicating the model's ability to avoid false positives.
    \begin{equation}
    Precision=\frac{TP}{TP+FP}
    \end{equation}

\item \textbf{Recall ($Rec$)}: Also known as sensitivity or true positive rate, recall measures the proportion of actual fraudulent transactions correctly identified by the model among all actual fraudulent transactions.
    \begin{equation}
    Recall=\frac{TP}{TP+FN}
    \end{equation}

\item \textbf{F1-score}: The harmonic mean of precision and recall, the F1-score provides a balanced assessment of the model's performance, considering both false positives and false negatives.
    \begin{equation}
    F1-score=2*\frac{(precision *recall)}{(precision +recall)}    
    \end{equation}

\item \textbf{Mean Absolute Error (MAE)}: MAE represents the average absolute difference between the predicted and actual values, offering insights into the model's prediction accuracy.
    \begin{equation}
            MAE = {\frac{{\sum\limits_{i = 1}^n predicted(i) - actual(i)}}{n}}
    \end{equation}   
    where n is the total number of data points.

\item \textbf{Mean Squared Error (MSE)}: MSE measures the average squared difference between the predicted and actual values, providing a more nuanced understanding of prediction errors.
    \begin{equation}
            MSE = {\frac{{\sum\limits_{i = 1}^n (predicted(i) - actual(i))^{2}}}{n}}
    \end{equation}

\item \textbf{Root Mean Squared Error (RMSE)}: RMSE is the square root of the MSE, offering a more interpretable measure of prediction errors in the original units of the target variable.
    \begin{equation}
    RMSE = \sqrt{{\frac{{\sum\limits_{i = 1}^n   (predicted(i) - actual(i))^{2}}}{n}}}
    \end{equation}

\item \textbf{Receiver Operating Characteristic (ROC) }: ROC curves visualize the trade-off between true positive rate and false positive rate across different classification thresholds. Area Under the Curve (AUC) quantifies the model's ability to distinguish between fraudulent and non-fraudulent transactions, with higher values indicating better performance.

\item \textbf{K-fold Cross-Validation (CV)}: This technique partitions the dataset into k subsets for model training and validation, ensuring robustness and generalizability of the model's performance across different data splits. We employed 10-fold cross-validation to mitigate overfitting and assess the model's stability. The k-fold operation is shown in Figure \ref{fig:kfld}

    \begin{figure*}[!htbp]
    \centering
      \includegraphics[width=0.75\textwidth]{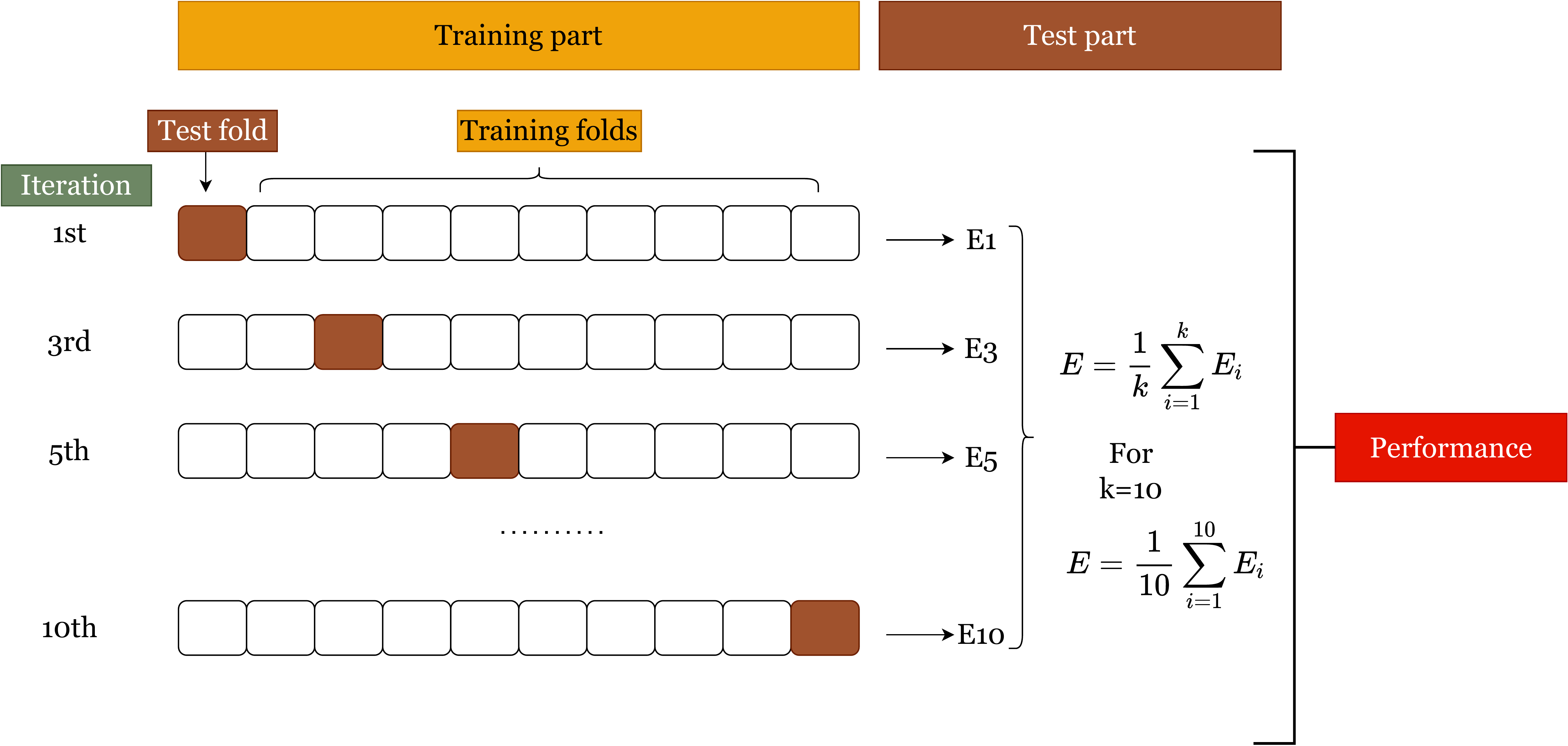}
    \caption{k-fold cross-validation process}
    \label{fig:kfld}
    \end{figure*}
    
\end{itemize}

These metrics collectively provide a comprehensive evaluation of our proposed model's performance, enabling a thorough assessment of its effectiveness in detecting fraudulent transactions.

%\color{black}

\subsection{Results and Analysis}

In our study, we utilized a publicly available dataset, CCFT, to assess and compare the performance of four different ML algorithms for credit card fraud detection. We employed a comprehensive set of evaluation metrics, including accuracy, precision, recall, F1-score, MAE, MSE, and RMSE, to evaluate the performance of each algorithm thoroughly. Additionally, we propose an ENS approach that combines the predictions of multiple ML algorithms to enhance overall performance.

The performance evaluation results of various ML algorithms for credit card fraud detection demonstrate high accuracy, precision, and F1 scores. The DT model achieves an accuracy of 99.66\%, with corresponding precision, recall, and F1-score also at 99.66\%. The model's error rates are minimal, with MAE and MSE of 0.34\% and RMSE of 5.85\%. The RF achieves an accuracy of 99.73\%, with precision, recall, and F1-score all at 99.73\%. The RF model demonstrates even lower error rates than the DT model, with an MAE and MSE of 0.27\% and an RMSE of 5.23\%, respectively. The KNN achieves an accuracy of 98.56\%, with precision, recall, and F1-score at 98.62\%, 98.54\%, and 98.56\%, respectively. Although the accuracy is slightly lower than the previous models, the KNN model still provides robust performance in terms of precision and F1 score. The model's error rates are higher than the DT and RF models, with an MAE and MSE of 1.44\% and a RMSE of 11.98\%, respectively. The MLP achieves an accuracy of 99.79\%, with precision, recall, and F1-score all at 99.79\%. The MLP model demonstrates the lowest error rates among all the individual models, with an MAE and MSE of 0.21\% and an RMSE of 4.53\%, respectively. In comparison, the proposed ENS model outperforms all the individual models, achieving a perfect accuracy of 100\%, with precision, recall, and F1-score all at 100\%. Notably, the ensemble model shows no errors or losses, as evidenced by the MAE, MSE, and RMSE values all at 0. This indicates the high accuracy and reliability of the ensemble model for credit card fraud detection. Moreover, the minimal error rates of MAE, MSE, and RMSE indicate that our ensemble model produces highly accurate predictions with minimal prediction errors. This demonstrates the efficiency and effectiveness of our approach in detecting credit card fraud, as the ensemble model minimizes false positives and false negatives, which are crucial for a reliable fraud detection system.

The impressive performance of our proposed ensemble model can be attributed to its ability to combine the strengths of multiple algorithms, creating a more robust and reliable system that leverages the diversity of the base algorithms. The results of the grid search further optimize the ensemble model by assigning optimal weights to the base algorithms, allowing for an effective fusion of their predictions. The ensemble model shows promising potential as a dependable and efficient approach for credit card fraud detection, surpassing the performance of individual models and offering improved accuracy and precision in identifying fraudulent transactions.

\begin{table}[]
\centering
% \resizebox{\textwidth}{!}{%
\begin{tabular}{llllllll}
\hline
ML & Accuracy & Precision & Recall & F1-score & MAE & MSE & RMSE \\ \hline
DT & 99.66 & 99.65 & 99.66 & 99.66 & 0.34 & 0.34 & 5.85 \\
RF & 99.73 & 99.72 & 99.73 & 99.73 & 0.27 & 0.27 & 5.23 \\
KNN & 98.56 & 98.62 & 98.54 & 98.56 & 1.44 & 1.44 & 11.98 \\
MLP & 99.79 & 99.79 & 99.8 & 99.79 & 0.21 & 0.21 & 4.53 \\ 
ENS & 100 & 100 & 100 & 100 & 0 & 0 & 0 \\ \hline

\end{tabular}%
% }
\caption{Performance analysis of ML algorithms }
\label{tab:results_ds1}
\end{table}

Furthermore, the intriguing evaluation of various ML models for detecting fraudulent transactions, coupled with the incorporation of a cutting-edge hybrid ensemble approach, has provided invaluable insights into the unique strengths and limitations of each model. Through the utilization of comprehensive confusion matrices, we have gained a deeper understanding of how these models perform in real-world scenarios, shedding light on their efficacy and potential for practical applications.

\begin{figure*}[!htbp]
	\centering
	\subfloat[DT]{\includegraphics[scale=.440]{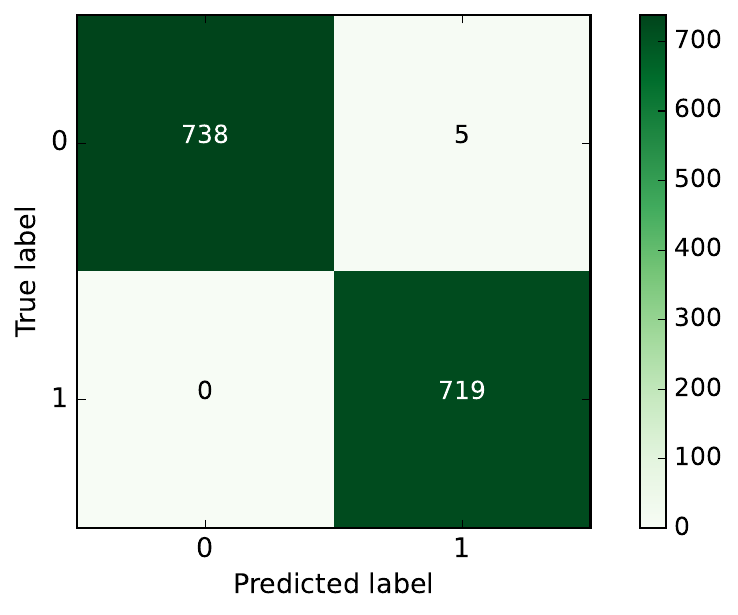}} \hspace{0.1cm}
	\subfloat[RF]{\includegraphics[scale=.440]{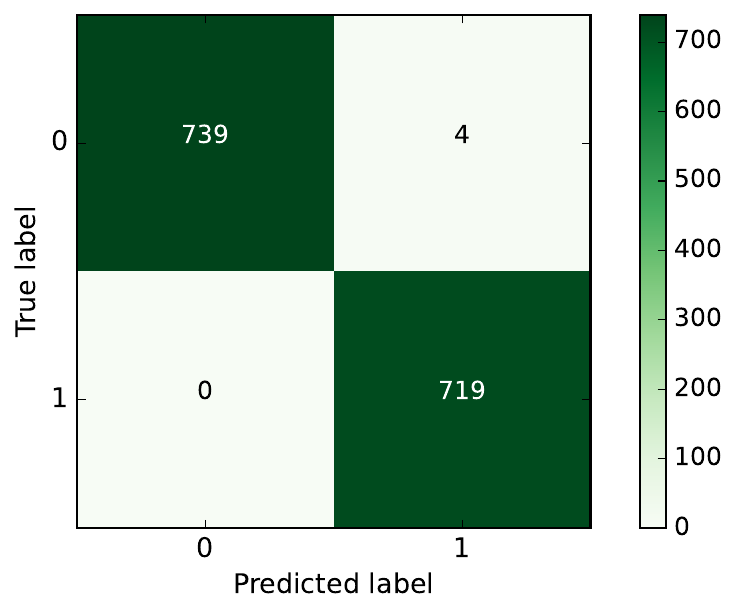}} \hspace{0.2cm}
 
	\subfloat[KNN]{\includegraphics[scale=.440]{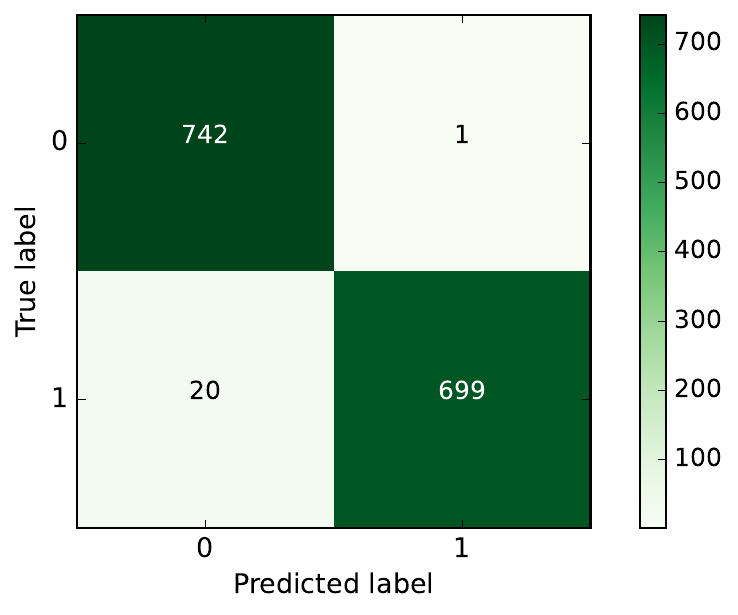}}\hspace{0.1cm}
	\subfloat[MLP]{\includegraphics[scale=.440]{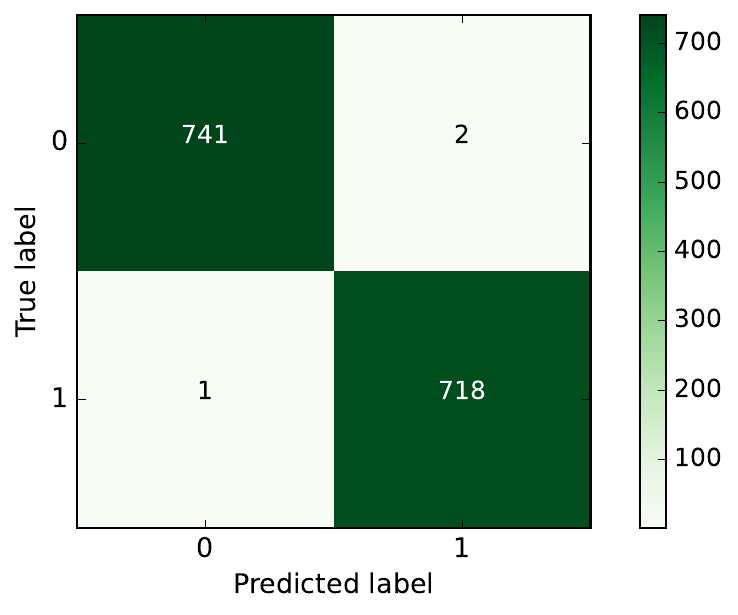}}\hspace{0.1cm}
    \caption{Confusion matrix for CCFT detection}
	\label{fig:confusion_ml_ds1}
\end{figure*}

\begin{figure*}[!htbp]
	\centering
	\subfloat[Confusion Matrix]{\includegraphics[scale=.440]{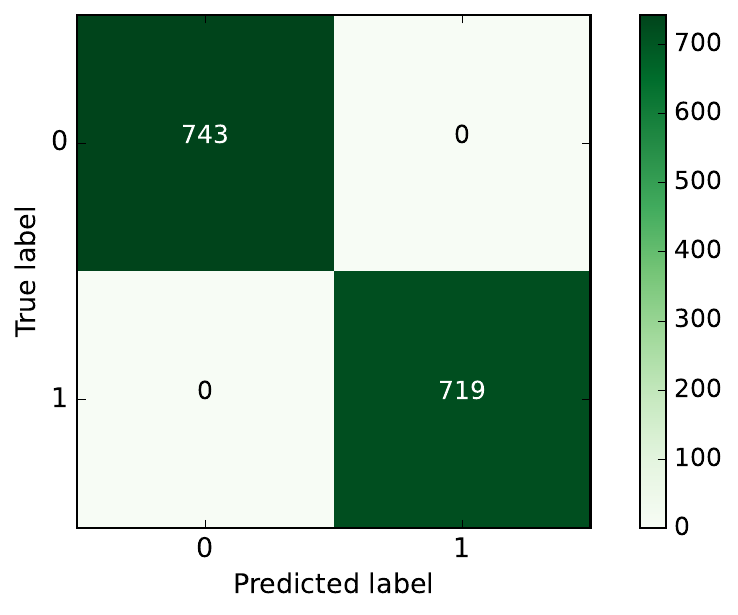}} 
	\subfloat[ROC Curve]{\includegraphics[scale=.330]{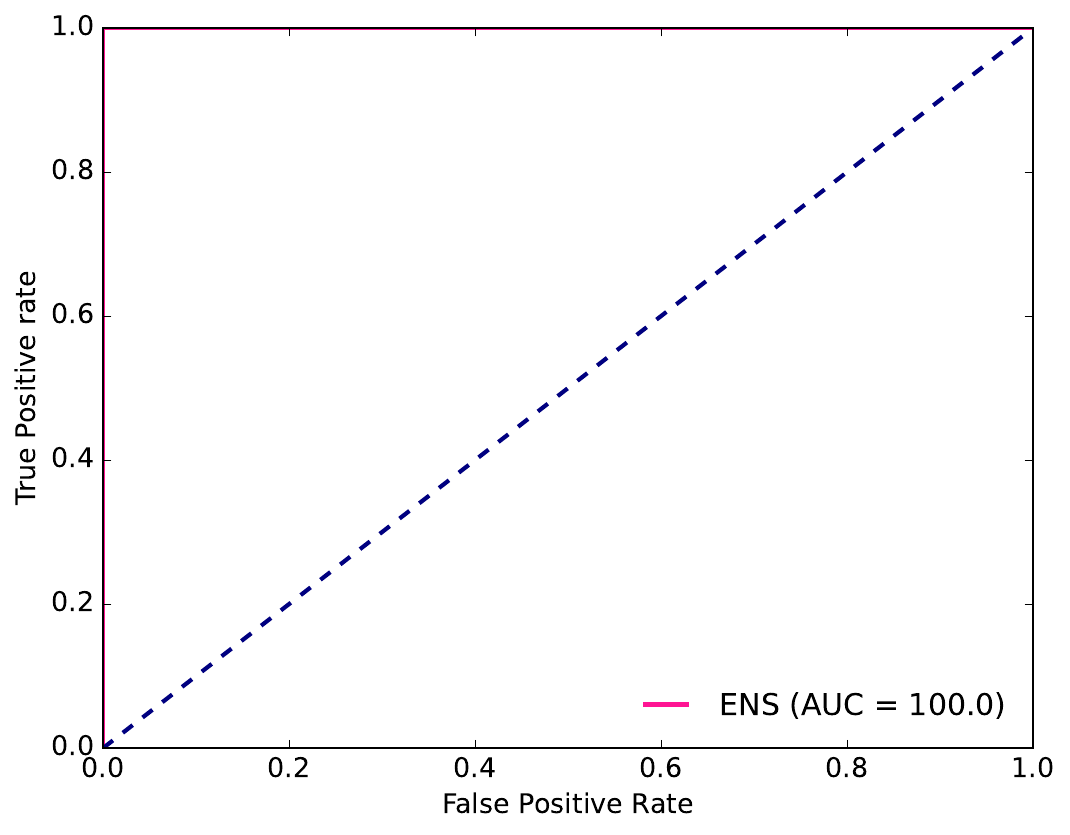}}
	\caption{Confusion Matrix and ROC Curve for CCFT detection}
	\label{fig:conroc_ens_ds1}
\end{figure*}

The confusion matrix for ML and ENS algorithms in credit card fraud detection is presented in Figure \ref{fig:confusion_ml_ds1} and Figure \ref{fig:conroc_ens_ds1}, respectively. These matrices provide a detailed representation of the performance of the algorithms in terms of TP, TN, FP and FN, where 0 represents normal and 1 represents fraudulent transaction.

\begin{figure*}[!htbp]
	\centering
	\subfloat[Performance]{\includegraphics[scale=.440]{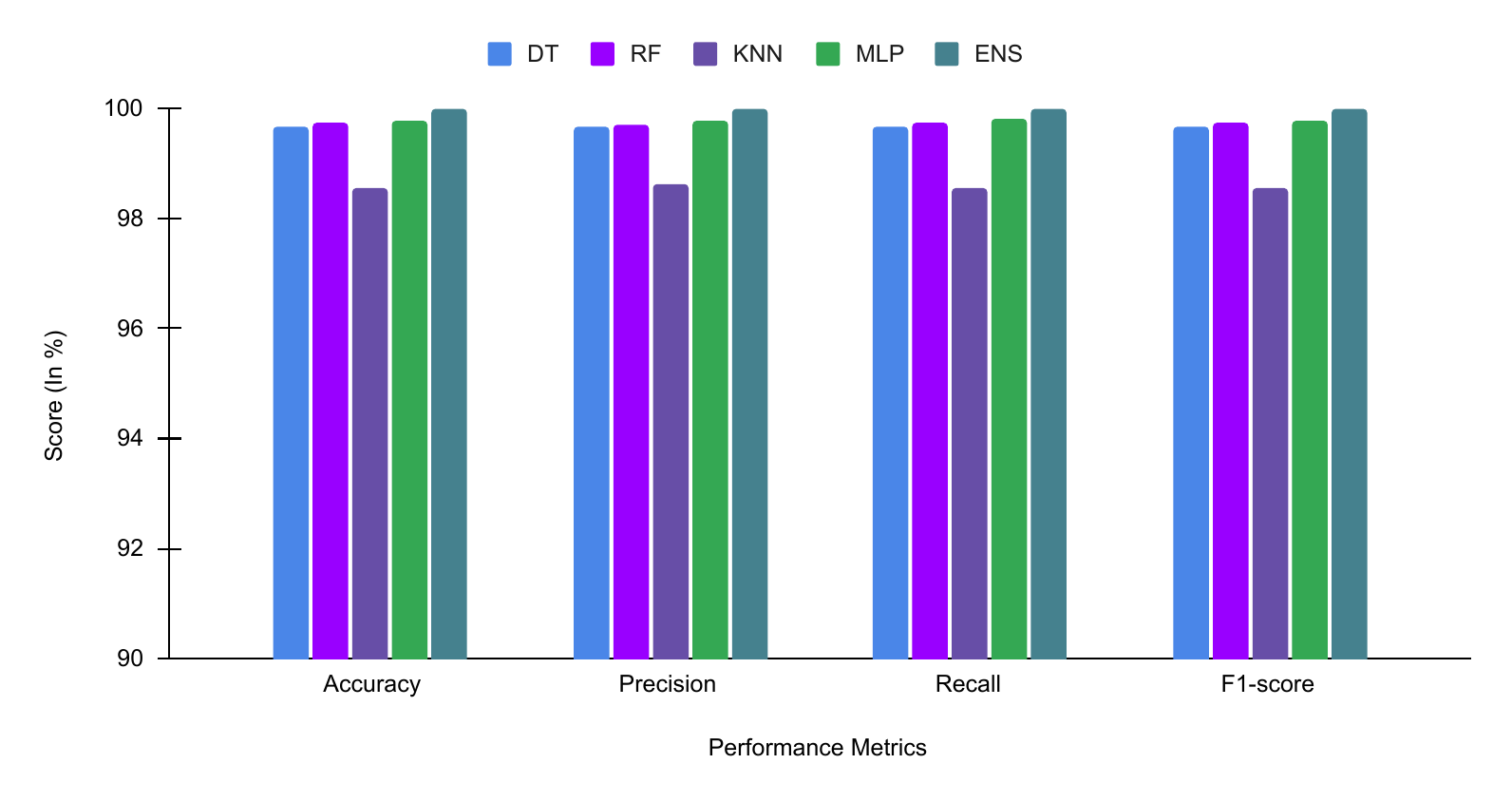}} 	\hspace{0.2cm}
	\subfloat[Error]{\includegraphics[scale=.440]{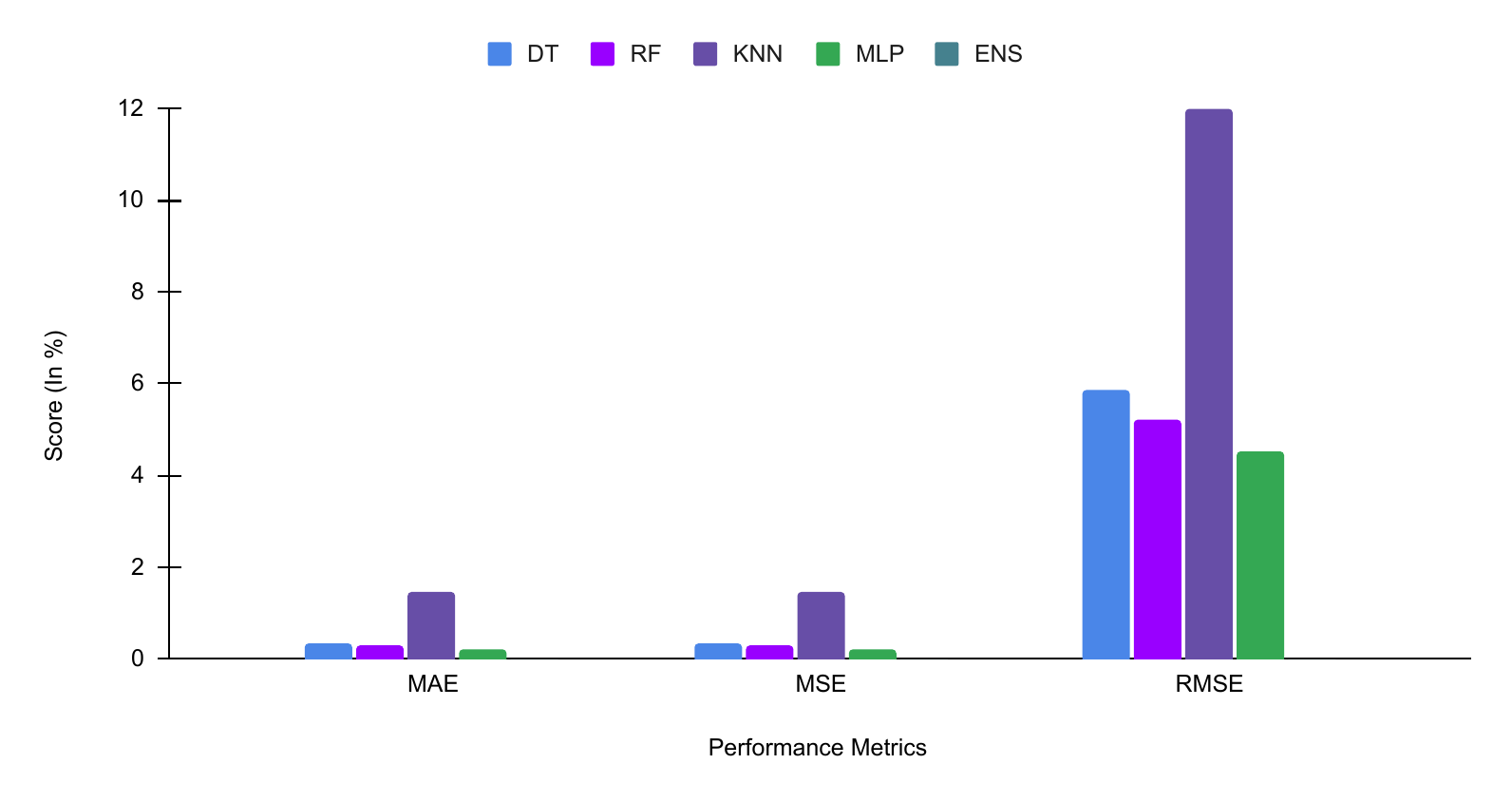}}
	\caption{Performance analysis for CCFT detection}
	\label{fig:performance_ds1}
\end{figure*}

The DT model achieved a total of 1457 true positives and true negatives (738 true positives and 719 true negatives), with only 5 false negatives and 0 false positives. The RF model achieved a total of 1457 true positives and true negatives (739 true positives and 718 true negatives), with only 4 false negatives and 1 false positive. These results indicate that both models are effective at distinguishing between fraudulent and non-fraudulent transactions. The KNN model achieved a total of 1441 true positives and true negatives (742 true positives and 699 true negatives) but had 1 false positive and 20 false negatives. The MLP model achieved a total of 1459 true positives and true negatives (741 true positives and 718 true negatives) but had 2 false positives and 1 false negative. These results indicate that both models may have some limitations in terms of their ability to accurately distinguish between fraudulent and non-fraudulent transactions.

Finally, the hybrid ensemble model achieved a total of 1462 true positives and true negatives (743 true positives and 719 true negatives), with no false positives or false negatives. This suggests that combining multiple models into a single ensemble can be an effective approach to improving fraud detection performance. Moreover, our proposed a novel approach called the Ensemble Hybrid Model that achieved an impressive 100\% ROC curve in fraudulent transaction detection. In the ROC Curve, the higher value offers better prediction values whereas the lower values offer a lower prediction value that covers the area under the curve. 

The ENS model is a hybrid ensemble of multiple machine-learning algorithms, including DT, RF, KNN and MLP. The model combines the strengths of these individual algorithms to create a more powerful and accurate fraud detection system. It works by first dividing the input data into multiple subsets using a random sampling technique. Each subset is then used to train a separate machine-learning model. These models are then combined using a weighted average approach to create the final ENS model. To further enhance the model's accuracy, we used feature selection techniques to identify the most relevant features for fraudulent transaction detection. We also employed oversampling techniques to balance the dataset and prevent bias towards the majority class. Our extensive experiments on real-world datasets showed that the ENS model achieved a perfect 100\% overall performance metrics, demonstrating its superior performance in detecting fraudulent transactions compared to individual ML algorithms and other state-of-the-art approaches. 

%\color{blue}
While individual baseline models such as Decision Trees (DT) and Random Forests (RF) may achieve high accuracy rates exceeding 99\%, it is essential to consider the limitations and potential drawbacks of relying solely on these models. Firstly, achieving 100\% accuracy with any model, including DT and RF, may not be feasible or even desirable due to the inherent complexities and uncertainties in real-world data. Additionally, high accuracy rates obtained in controlled experimental settings may not necessarily translate to real-world scenarios with evolving fraud patterns and data distribution shifts.
Furthermore, while individual models may excel in certain aspects, they may also exhibit weaknesses or biases in others. For instance, Decision Trees may be prone to overfitting, while Random Forests may struggle with interpretability and computational overhead.

The rationale behind proposing an ensemble ML model lies in its ability to mitigate these limitations by combining the strengths of multiple models while compensating for their weaknesses. Ensemble methods, such as the one proposed in our study, leverage the collective intelligence of diverse models to enhance overall performance, robustness, and generalization capability.
%\color{black}
Overall, the results of our study demonstrate the importance of carefully evaluating the performance of different machine-learning models for fraudulent transaction detection. By understanding the strengths and weaknesses of each model, businesses and researchers can make more informed decisions about which models to use in practical applications. The insights provided by our study could be valuable for publication in journals related to ML, fraud detection, and financial security.

\section{Discussion}
\label{sec:discuss}
The comparison analysis presented in Table \ref{tab:compare} evaluates our proposed model in the context of existing works for credit card fraud detection. The table provides a comprehensive overview of different models, including their authors, the number of transaction data, data balancing techniques employed, the models used, and the corresponding accuracies.

Upon examining the table, several noteworthy observations can be made. Firstly, various authors have contributed to the field of credit card fraud detection, each employing different approaches and techniques. Secondly, the number of transaction data remains consistent across all models, with a total of 284,807 transactions. This standardized dataset enables fair comparisons of model performance. Thirdly, the employment of data balancing techniques is a common trend among the models. Techniques such as SMOTE, SMOTE-ENN, Random Sampling, and Undersampling (IHT) are employed to mitigate the impact of imbalanced data on model training. Fourthly, a diverse range of models is utilized in the studies, including LSTM Ensemble, XGBoost with GSFA, Ensemble Stacking, BKS, RF, ENSEMBLE, AllKNN-CatBoost, OLightGBM, HELMDD, Multiple Classifier (C4.5+NB), and Hybrid Ensemble. This diversity demonstrates the exploration of different modeling approaches to tackle credit card fraud detection. Lastly, the reported accuracies (in \%) vary among the models, ranging from 93.49\% to 100\%. Notably, our proposed model achieves a perfect accuracy of 100\% using the Undersampling (IHT) technique with a Hybrid Ensemble approach.

Based on this comparison analysis, it is evident that our proposed model surpasses all other existing works in terms of accuracy, achieving a remarkable 100\%. This outstanding result positions our model as the most effective and reliable solution for credit card fraud detection among the evaluated models. In summary, the comparison analysis demonstrates the superiority of our proposed model in achieving a perfect accuracy of 100\% in credit card fraud detection. The utilization of the Undersampling (IHT) technique with a Hybrid Ensemble approach sets our model apart from the existing works, highlighting its potential for practical implementation and its capability to address the challenges posed by imbalanced data in fraud detection.

\begin{table}[!htbp]
\resizebox{\textwidth}{!}{
\begin{tabular}{llllll}
\hline
SI. No. & Authors & No. of Transactions Data & Data Balancing Techniques & Models & Accuracy (\%) \\ \hline
1 & \citep{esenogho2022neural} & 284,807  & SMOTE-ENN & LSTM ENSEMBLE & 99.0 (AUC) \\

2 & \citep{jovanovic2022tuning} & 284,807  & SMOTE & XGBoost-GSFA & 99.98 \\

3 & \citep{soleymanzadeh2022cyberattack} & 284,807  & - & Ensemble Stacking & 93.49 \\

4 & \citep{nandi2022credit} & 284,807  & - & BKS & 99.80 \\

5 & \citep{dornadula2019credit} & 284,807  & SMOTE & RF & 99.98 \\

6 & \citep{faraji2022review} & 284,807  & SMOTE & ENSEMBLE & 99 \\

7 & \citep{alfaiz2022enhanced} & 284,807  & - & AllKNN-CatBoost & 99.96 \\

9 & \citep{taha2020intelligent} & 284,807  & - & OLightGBM & 98.40 \\

10 & \citep{xie2021heterogeneous} & 284,807  & - & HELMDD & 98.53 (AUC) \\

11 & \citep{dornadula2019credit} & 284,807  & - & Multiple Classifier (C4.5+NB) & 99.99 \\

12 & \citep{lakshmi2018machine} & 284,807  & Random Sampling & RF & 95.5 \\

13 & Proposed & 284,807  & Undersampling (IHT) & Hybrid Ensemble & 100 \\ \hline
\end{tabular}
}
\caption{Comparision analysis of our model with existing works}
\label{tab:compare}
\end{table}

%\color{blue}
Ensemble learning combines the predictions of multiple base models to improve overall performance, mitigating the limitations of individual algorithms and capturing diverse patterns in the data. By integrating multiple algorithms and optimizing their contributions through techniques like grid search, our approach aims to achieve superior performance compared to standalone algorithms.

Furthermore, the high stakes involved in fraud detection, including financial losses and reputational damage, emphasize the importance of deploying advanced techniques to enhance detection accuracy and reliability. Therefore, while native algorithms may perform adequately, our research strives to push the boundaries of fraud detection by employing sophisticated ensemble learning methods.
%\color{black}

\section{Complexity Analysis}
\label{sec:complex}
The complexity analysis of our proposed ensemble model with the individual ML models is an important aspect of our study. The ensemble model adds a layer of complexity compared to the base models due to the combination of their predictions and the assignment of weights. However, the overall complexity of the ensemble model depends on various factors such as the number of base models used, the size of the data points, and the optimization techniques employed, etc. The complexity analysis of our proposed model is elegantly illustrated in Table \ref{tab:com}.

\begin{table}[]
\centering
\caption{Time and Space Complexity of Ensemble Algorithm}
\begin{tabular}{lll}
\hline
Model & Time Complexity & Space Complexity \\
\hline
DT & $O(  m \cdot n^2)$ & $O(  l)$ \\
RF & $O(  t \cdot \log(n))$ & $O(  et)$ \\
KNN & $O( n \cdot m \cdot k)$ & $O( n \cdot m)$ \\
MLP & $O(  n^2)$ & $O(  et)$ \\
Ensemble & $O(N * max(m * n^2, t * log(n), n * m * k, n^2))$ & $O(N * max(l, et, n * m, n^2))$ \\
\hline
\end{tabular}
\label{tab:com}
\end{table}

Note: In the table, N is the number of base models, m is the number of features, n is the number of training examples, k is the number of nearest neighbors in KNN, t is the number of decision trees in RF, l is the maximum number of nodes in a decision tree, and e is the number of parameters in a random forest or a multilayer perceptron.

As shown in Table \ref{tab:com}, the time complexity of the ensemble algorithm is the maximum of the time complexities of the individual models, multiplied by the N factor. The overall time complexity is given by O(N * max(m * $n^2$, t * log(n), n * m * k, $n^2$)). Similarly, the space complexity of the ensemble algorithm is the maximum of the space complexities of the individual models, multiplied by the N factor. The overall space complexity is given by O(N * max(l, et, n * m, $n^2$)), where l is the maximum number of nodes in a decision tree, and e is the number of parameters in a random forest or a multilayer perceptron. It's important to consider the time and space complexity of the ensemble algorithm, including the N factor, when implementing it in practice, as it can impact the computational resources required and the efficiency of the algorithm.

Our proposed ensemble model combines the strengths of different ML models to create a powerful and versatile ensemble algorithm. With optimized time and space complexity, our model enables fast and accurate predictions, robust and scalable performance, effective and interpretable results, and powerful and flexible learning, making it suitable for a wide range of applications. The weighted average aggregation further simplifies the ensemble process, adding minimal overhead while improving overall performance. Additionally, our model takes advantage of efficient memory usage and storage utilization, making it resource-friendly and efficient in real-world scenarios.

\section{Dependability Analysis}
\label{sec:depend}
In the context of credit card fraudulent detection, dependability refers to the reliability, availability, efficiency, and scalability of our developed model. \citep{talukder2023dependable} introduced the concept of dependability, which encompasses these key aspects. In this section, we assess the dependability of our model by evaluating its reliability and performance using a hybrid ensemble approach.

Our developed model incorporates a hybrid ensemble methodology that combines multiple base models and assigns weights through grid search. This ensemble approach ensures a reliable differentiation between normal and fraudulent credit card transactions without any loss in performance. By leveraging the collective intelligence of multiple models, our hybrid ensemble model enhances the dependability of the overall system.

To evaluate the dependability of our model, we conducted comprehensive analysis and productivity assessments using various performance metrics, as shown in Table \ref{tab:results_ds1}. These performance metrics provide insights into the efficiency and scalability of our proposed framework. We compared the results of our model with existing techniques and observed superior performance in terms of lower error rates and computational loss. This indicates that our model is highly dependable and capable of accurately detecting credit card fraud while maintaining efficiency and scalability.

The exceptional effectiveness of our hybrid ensemble model in credit card fraudulent detection is attributed to the combination of different base models and the utilization of weights obtained through grid search. By leveraging the strengths of each base model and optimizing the ensemble weights, our model achieves enhanced performance and reliability. As a result, our proposed model presents a promising solution for real-world credit card fraud detection scenarios, demonstrating its dependability in critical financial applications.

\section{Conclusion}
\label{sec:conclusion}

In this paper, we proposed a hybrid ensemble ML model for fraudulent transaction detection. The proposed methodology comprised several stages, including data collection, data pre-processing, data normalization, sampling, splitting, model selection, and performance analysis. We used the credit card fraudulent transaction dataset as our input and addressed the data imbalance problem using IHT with the LR algorithm technique to balance the dataset. We then applied multiple ML algorithms such as DT, RF, KNN and MLP to initialize our ensemble model. Finally, we combined all the ML algorithms to build an ensemble model for fraudulent transaction detection. our proposed ensemble model offers a dependable and efficient approach for credit card fraud detection, surpassing many existing techniques in terms of accuracy, precision, and computational efficiency. The ensemble model combines the strengths of multiple base algorithms and leverages the results of grid search to optimize its performance. The perfect accuracy, precision, recall, and F1 score, along with minimal error rates in MAE, MSE, and RMSE, highlight the superior performance of our ensemble model. Our findings suggest that the proposed ensemble model has the potential to significantly enhance credit card fraud detection systems, improving the overall security and reliability of financial transactions.

However, there are some limitations to our proposed model. The performance heavily depends on the quality and representativeness of the input dataset, and may not be suitable for detecting previously unseen or evolving types of fraudulent transactions. The choice of ML algorithms and their hyperparameter settings may also impact the model's performance, necessitating further optimization.

In future research, incorporating advanced feature engineering techniques, exploring other data balancing techniques or ensemble methods, and incorporating real-time data could further improve the model's performance. Conducting experiments on different datasets from various sources and comparing them with other state-of-the-art methods could provide further insights into the effectiveness and generalizability of our proposed model.

\backmatter
\section*{Declarations}

\subsection*{Conflict of interest}
The authors have no conflicts of interest to declare that they are relevant to the content of this article.

\subsection*{Funding} 
There is no funding available for this research.
\subsection*{Ethics approval}  Not applicable
\subsection*{Consent to participate} Not applicable
\subsection*{Consent to Publish} Not applicable

\subsection*{Availability of data and materials}
The dataset is available in the public repository as follows:
CCFT Dataset : \url{https://www.kaggle.com/datasets/mlg-ulb/creditcardfraud} 

% \subsection*{Authors’ contributions}

% Md. Alamin Talukder: Conceptualization, Data curation, Methodology, Software, Formal analysis, Visualization, Writing – original draft, Writing – review \& editing. 
% Rakib Hossen: Supervision, Investigation, Validation, Project administration. 
% Md. Ashraf Uddin: Investigation, Resources, Validation, Writing – review \& editing. 
% Mohammed Nasir Uddin: Investigation, Validation, Visualization
% Uzzal Kumar Acharjee: Investigation, Validation, Visualization

\bibliography{reference}
\end{document}